\def\year{2021}\relax
\documentclass[letterpaper]{article} 
\usepackage{aaai21}  
\usepackage{times}  
\usepackage{helvet} 
\usepackage{courier}  
\usepackage[hyphens]{url}  
\usepackage{graphicx} 
\usepackage{natbib}  
\usepackage{caption} 
\def\eg{\emph{e.g}.}

\usepackage{amsmath}  
\usepackage{multirow}
\usepackage{docmute}
\usepackage{tabularx}

\newcommand{\equref}[1]{(\ref{#1})}
\newcommand{\figref}[1]{Fig. \ref{#1}}
\newcommand{\tabref}[1]{Table \ref{#1}}
\newcommand{\secref}[1]{Section \ref{#1}}
\renewcommand\footnotemark{}

\usepackage{array, boldline, makecell, booktabs}
\usepackage{subfigure}
\usepackage{amssymb}
\usepackage[switch]{lineno}  %

\title{Cross-Domain Grouping and Alignment\\for Domain Adaptive Semantic Segmentation}
 \author {
         Minsu Kim\textsuperscript{\rm 1},
         Sunghun Joung\textsuperscript{\rm 1},
         Seungryong Kim\textsuperscript{\rm 2},
         Jungin Park\textsuperscript{\rm 1},
         Ig-Jae Kim\textsuperscript{\rm 3}, 
         Kwanghoon Sohn\textsuperscript{\rm 1,*}\thanks{$^{*}$Corresponding author} 
         \\
 }

 \affiliations {
     \textsuperscript{\rm 1} Yonsei University \textsuperscript{\rm 2} Korea University \textsuperscript{\rm 3} Korea Institute of Science and Technology (KIST)\\
     \{minsukim320, sunghunjoung, newrun, khsohn\}@yonsei.ac.kr, seungryong\_kim@korea.ac.kr,
     drjay@kist.re.kr
 }

\begin{document}
\maketitle

\begin{abstract}
Existing techniques to adapt semantic segmentation networks across source and target domains within deep convolutional neural networks (CNNs) deal with all the samples from the two domains in a global or category-aware manner. They do not consider an inter-class variation within the target domain itself or estimated category, providing the limitation to encode the domains having a multi-modal data distribution. To overcome this limitation, we introduce a learnable clustering module, and a novel domain adaptation framework, called cross-domain grouping and alignment. To cluster the samples across domains with an aim to maximize the domain alignment without forgetting precise segmentation ability on the source domain, we present two loss functions, in particular, for encouraging semantic consistency and orthogonality among the clusters. We also present a loss so as to solve a class imbalance problem, which is the other limitation of the previous methods. Our experiments show that our method consistently boosts the adaptation performance in semantic segmentation, outperforming the state-of-the-arts on various domain adaptation settings.

	\end{abstract}

\section{Introduction}\label{sec:1}
Semantic segmentation aims at densely assigning semantic category label to each pixel given an image.
Though the remarkable progresses have been dominated by deep neural networks trained on large-scale labeled dataset~\cite{Liang17}.
The segmentation model trained on the labeled data in source domain usually cannot generalize well to the unseen data in target domain.
For example, the model trained on the data from one city or computer-generated scene~\cite{Stephan16,German16} may fail to yield accurate pixel-level predictions for the scenes of another city or real scene.
The main reason lies in the different data distribution between such source and target domains, typically known as domain discrepancy~\cite{Hidetoshi00}.

\begin{figure}[t]
	\centering
	\includegraphics[width=0.99 \linewidth]{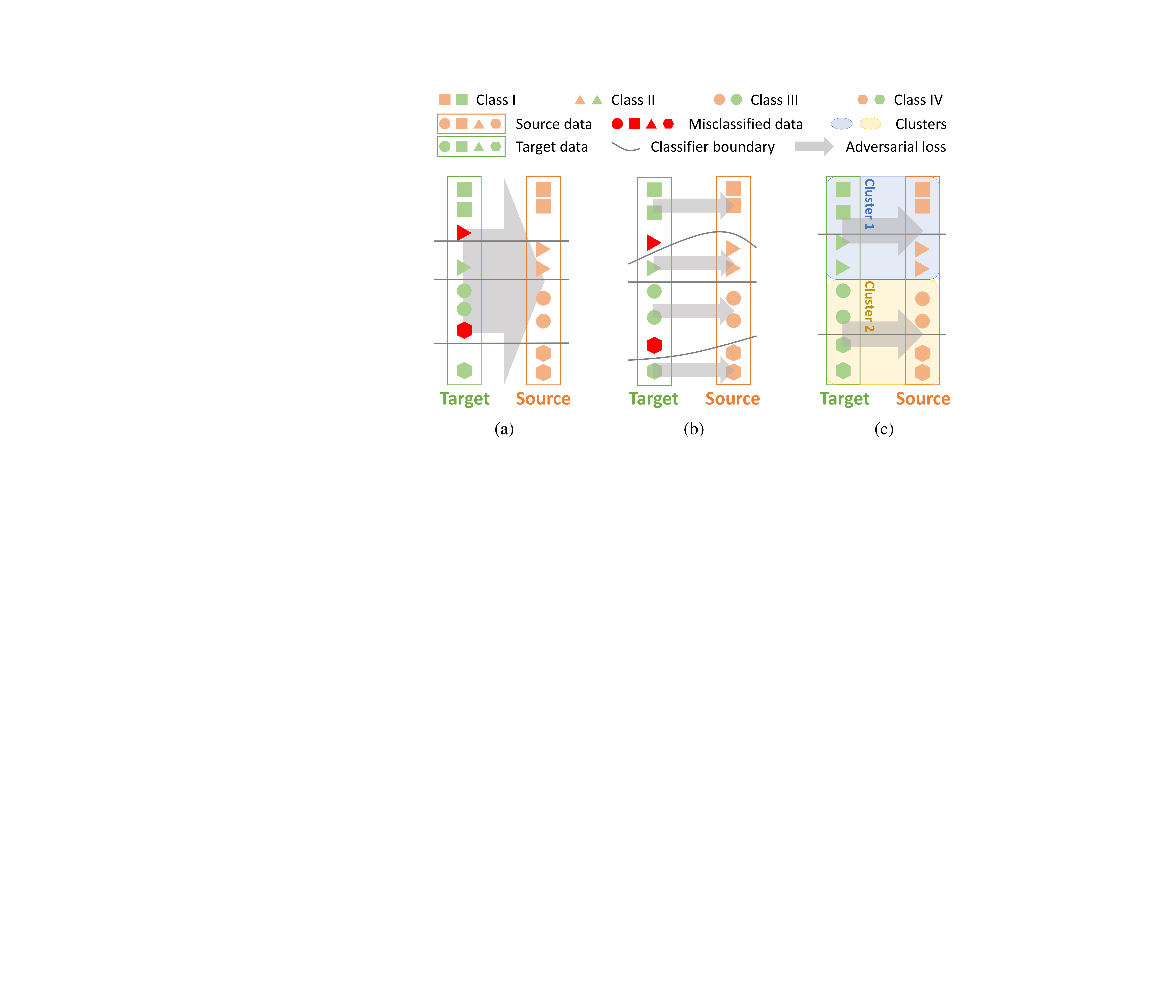}
	\caption{Illustration of cross-domain grouping and alignment :  Conventional methods aim to reduce the domain discrepancy between source and target domains through (a) global and (b) category-level domain alignment, without taking into account the \emph{inter-class} variation or rely solely on the category classifier. (c) We propose to replace this category classifier with an intermediate cross-domain grouping module to align each group separately (best view in color).}
	\label{fig:f1}
\end{figure}

To address this issue, domain adaptive semantic segmentation methods have been proposed in which they align data distribution between the source and target domains by adopting a domain discriminator~\cite{Judy16,Yi18}.
Formally, these methods aim to minimize an adversarial loss~\cite{Ian14} to reduce the domain discrepancy at image-level~\cite{Zuxuan18,Judy18,Chang19}, feature-level~\cite{Judy16}, and category probability-level~\cite{Yang18,Yunsheng19,Yi18} distributions without forgetting semantic segmentation ability on the source domain.
However, their accuracy is still limited when aligning multi-modal data distribution~\cite{Sanjeev17}, which cannot guarantee that the target samples from different categories are properly separated as in \figref{fig:f1} (a).


To tackle this limitation, category-level domain adaptation methods ~\cite{Yi17,Liang19} have been proposed for semantic segmentation in which they minimize the class-specific domain discrepancy across the source and target domains.
Together with supervision from the source domain, this enforces the segmentation network to learn discriminative representation for different classes on both domains.
They utilize a category classifier trained on the source domain to generate \emph{pseudo} class labels on the target domain.
It results in inaccurate labels for domain adaptation that misleads the domain alignment and accumulates errors as in \figref{fig:f1} (b).
It also contains a class imbalance problem \cite{Yang18}, where the network works well for majority categories with a large number of pixels (e.g. road and building), while not suitable for minority categories with a small number of pixels (e.g. traffic sign).


To overcome this limitation, we present cross-domain grouping and alignment for domain adaptive semantic segmentation.
As illustrated in \figref{fig:f1} (c), the key idea of our method is to apply an intermediate grouping module to replace the category classifier,
allowing to align samples of source and target domains at each group to be similar without using error-prone category classifier.
To make the grouping module help with domain adaptation, we propose several losses in a manner that the category distribution of each group between different domains should be consistent, while the category distribution of different groups in the same domain should be orthogonal.
Furthermore, we present a group-level class equivalence scheme in order to align all the categories regardless of the number of pixels.
The proposed method is extensively evaluated through an ablation study and comparison with state-of-the-art methods on various domain adaptive semantic segmentation benchmarks, including GTA5 $\rightarrow$ Cityscapes and SYNTHIA $\rightarrow$ Cityscapes.

\section{Related Work}\label{sec:2}

\subsection{Semantic Segmentation} 
Numerous methods have been proposed to assign class labels in pixel level for input images.
Long et al.~\shortcite{Jonathan15} first transformed a classification convolutional neural network (CNN)~\cite{Alex12,Karen15,He16} to a fully-convolutional network (FCN) for semantic segmentation.
Following the line of FCN-based methods, several methods utilized dilated convolutions to enlarge the receptive field~\cite{Fisher15} and reason about spatial relationship~\cite{Liang17}.
Recently, Zhao et al.~\shortcite{Hengshuang17} presented pyramid pooling module to encode the global and local context.
Although these methods yielded impressive results in semantic segmentation, they still relied on large datasets with dense pixel-level class labels, which is expensive and laborious.
An alternative is to utilize synthetic data~\cite{Stephan16,German16} which can make unlimited amounts of labels available.
Nevertheless, synthetic data still suffer from a substantially different data distribution from real data, which results in a dramatic performance drop when applying the trained model to real scenes.

\subsection{Domain Adaptive Semantic Segmentation} 
Due to the obvious mismatch between synthetic and real data, unsupervised domain adaptation (UDA) is studied to minimize the domain discrepancy by aligning the feature distribution between source and target data.
As a pioneering work, Ganin et al.~\shortcite{Yaroslav15_1} introduced the domain adversarial network to transfer the feature distribution, and Tzeng et al.~\shortcite{Eric17} proposed adversarial discriminative alignment.

For pixel-level classification, numerous approaches~\cite{Zuxuan18,Judy18,Chang19} utilized image-level adaptation methods which translate source image to have the texture appearance of target image, while preserving the structure information of the source image for adapting cross-domain knowledge.
In contrast, several methods~\cite{Yang18,Yunsheng19,Guangrui20} adopted the iterative self-training approach to alternatively select unlabelled target samples with higher class probability and utilized them as a pseudo ground-truth.
The feature-level adaptation methods align the intermediate feature distribution via adversarial framework.
Hoffman et al.~\shortcite{Judy16} introduced a feature-level adaptation method to align the intermediate feature distribution for the global and local alignment.
Tsai et al.~\shortcite{Yi18} adopted output-level adaptation for structured output space, since it contains similar spatial structure with semantic segmentation.
However, these methods aim to align overall data distribution without taking into account the inter-class variation.

To solve this problem, several methods~\cite{Yi17,Liang19} introduced category-level adversarial learning to align the data distributions independently for each class.
Similarly, other works~\cite{Yi19, Jiaxing20} discovered patch-level adaptation methods by using multiple modes of patch-wise output distribution to differentiate the feature representation of patches.
However, inaccurate domain alignment occurs because these methods rely heavily on category or patch classifiers trained in the source domain.
The most similar to our work is Wang et al.~\shortcite{Zhonghao20}, which group the category classes into several groups for domain adaptive semantic segmentation.
While they divide stuff and things (i.e. disconnected regions), our cross-domain grouping module divides the categories into multiple groups that the grouping network and segmentation network can be trained in a joint and boosting manner.

\subsection{Unsupervised Deep Clustering}
\begin{figure*}
   \centering
   \includegraphics[width=0.99 \linewidth]{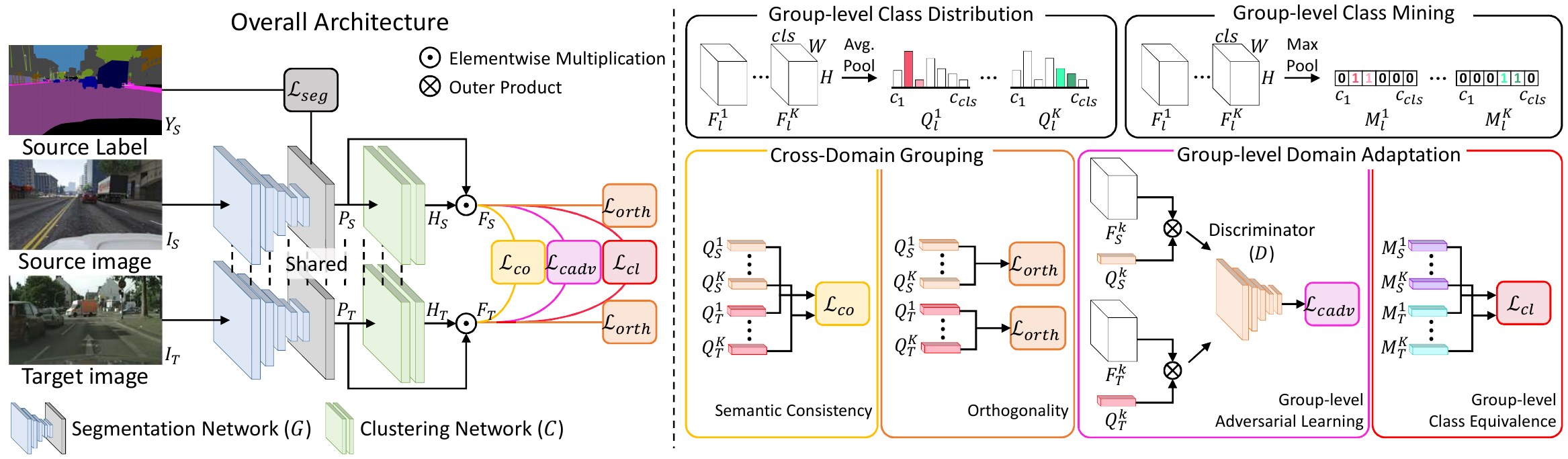}
   
   \caption{Overview of our method. Images from the source and target domains are passed through segmentation network $G$.
   We decompose the data distribution of source and target domains into a set of $K$ sub-spaces with cross-domain grouping network $C$. Then discriminator $D$ distinguishes whether the data distribution for each sub-space is from the source or target domain. 
   }
   
\label{fig:f2}
\end{figure*}
A variety of approaches have applied deep clustering algorithms that simultaneously discover groups in training data and perform representation learning.
Chang et al.~\shortcite{Jianlong17} proposed to cast the clustering problem into pairwise classification using CNN.
Caron et al.~\shortcite{Mathilde18} proposed a learning procedure that alternates between clustered images in the representation space and trains a model that assigns images to their clusters.
Other approaches~\cite{Armand12,Zhiqiang17} localized the salient and common objects by clustering pixels in multiple images.
Similarly, Collins et al.~\shortcite{Edo18} proposed deep feature factorization (DFF) to group the common part segments between images through non-negative matrix factorization (NMF)~\cite{chris05} on CNN features.
This paper follows such a strategy to group semantic consistent data representation across the source and target domains.


\section{Proposed Method}\label{sec:3}

\subsection{Problem Statement and Overview}\label{sec:31}
Let us denote the source and target images as ${I}_{S},{I}_{T}$
, where only the source data is annotated with per-pixel semantic categories as ${Y}_{S}$.
We seek to train a semantic segmentation network $G$, which outputs pixel-wise class probability $P_{S}, P_{T}$
on both source and target domains reliably, with height $h$, width $w$, and the number of classes $cls$, respectively.
Our goal is to train the segmentation network that yields to
align probability distribution of the source and target domains $P_{S}$ and $P_{T}$ so that the network $G$ can correctly predict the pixel-level labels even for the target data ${I}_{T}$, following recent study~\cite{Yi18} of adaptation in the output probability space, which shows better performance than adaptation in the intermediate feature space.

Conventionally two types of domain adaptation approaches have been proposed: 
\emph{global} domain adaptation and \emph{category-level} domain adaptation.
The former aims to align the global domain differences, while the latter aims to minimize class specific domain discrepancy for each category.
However, 
\emph{global} domain adaptation does not take into account the inter-class variations, and \emph{category-level} domain adaptation rely solely on category classifier.
To this end, we propose a novel method by clustering the samples as $K$ groups across the source and target domains.
Concretely, we cluster the probability distribution into $K$ groups using cross-domain grouping module, followed by \emph{group-level} domain alignment. 
By setting $K$ greater than 1, domain alignment of complicated data distribution can be solved by an alignment of $K$ simple data distributions which is the challenge in \emph{global} domain adaptation.
By setting $K$ less than $cls$, the domain misalignment in \emph{category-level} can be mitigated without using a category classifier trained in the source domain.
In the following, we introduce our overall network architecture (\secref{sec:32}), several constraints for cross-domain grouping (\secref{sec:33}), and cross-domain alignment (\secref{sec:34}).

\subsection{Network Architecture}\label{sec:32}

\figref{fig:f2} illustrates our overall framework.
Our network consists of three major components: 1) the semantic segmentation network $G$, 2) the cross-domain grouping module $C$ to cluster sub-spaces based on the output probability distribution, and 3) the discriminator $D$ for group-level domain adaptation.
In the following sections, we denote source and target domains as $l
\in\left\{S,T\right\}$ unless otherwise stated.

\begin{figure*}
   \centering
   \includegraphics[width=1 \linewidth]{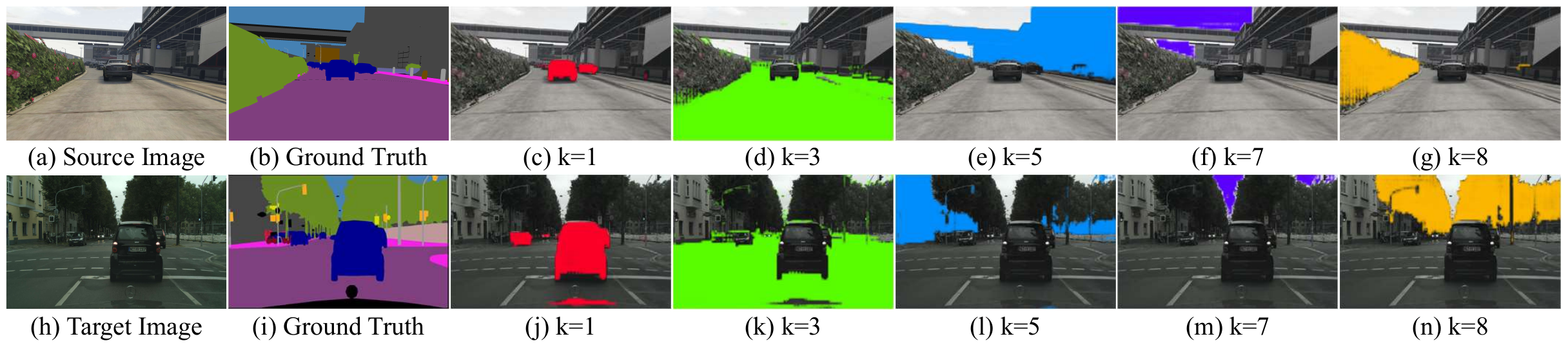}
   
   \caption{Visualization of cross-domain grouping on the source (first row) and target (second row) image with $K=8$. (From left to right) Input image and clustering results. Note that color represents the $K$ different sub-space.
   }
   \label{fig:NMF}
\end{figure*}

   

\subsubsection{Segmentation network.}
Following the works~\cite{Yi18,Yunsheng19,Zhonghao20}, we exploit DeepLab-V2~\shortcite{Liang17} with ResNet-101~\shortcite{He16} pre-trained on ImageNet~\shortcite{Jia09} dataset.
The source and target images ${I}_{l}$ are fed into the segmentation network $G$, outputting pixel-wise class probability distribution ${P}_{l}=G({I}_{l})$. 
Note that ${P}_{l}$ is extracted from the segmentation network before applying a softmax layer with same resolution as the input using bilinear interpolation, similar to Tsai et al.~\shortcite{Yi18}.

\subsubsection{Cross-Domain grouping network.}
Our cross-domain grouping network $C$ is formulated as two convolutions.
We design each convolution with $1\times1$ kernel and group mapping function.
The first convolution produces 64-channel feature, followed by ReLU and batch normalization.
The second convolution produces $K$ grouping scores, followed by softmax function to output group probability ${H}^{k}_{l}=C({P}_{l})$.
We then apply element-wise multiplication between ${H}^{k}_{l}$ and each channel dimension in ${P}_{l}$, obtaining group-specific feature ${F}^{k}_{l}$.
The cross-domain grouping network can be easily replaced with other learnable clustering methods.

\subsubsection{Discriminator.}
For group-level domain alignment, we fed ${F}^{k}_{l}$ into the discriminator $D$.
Following Li et al.~\shortcite{Yunsheng19}, we set the discriminator using five $4\times 4$ convolutional layers of stride 2, where the number of channels is $\{ 64, 128, 256, 512, 1 \}$ to form the network.
We use a leaky ReLU~\shortcite{Andrew13} parameterized by 0.2 which is utilized for each convolutional layer except the last one.

\subsection{Losses for Cross-Domain Grouping}\label{sec:33}
Perhaps one of the most straightforward ways of grouping is to utilize existing clustering methods, \eg{} k-means~\cite{Adam12} or non-negative matrix factorization (NMF)~\cite{Edo18}.
These strategies, however, are not learnable, and thus, they cannot weave the advantages of category-level domain information.
Unlike these, we present a learnable clustering module with two loss functions to take advantage of the category-level domain adaptation methods.
We discuss in more detail about the effectiveness of our grouping compared to non-learnable models~\cite{Adam12,Edo18} in \secref{sec:analysis}.
In the following, we present each loss function in detail.

\subsubsection{Semantic consistency.}\label{sec:331}
Our first insight about grouping is that the category distribution of each group between the source and target domains has to be consistent so that the clustered group can benefit from the category-level domain adaptation method.
To this end, we first estimate the class distribution ${Q}^{k}_{l}$ by using average pooling layer on each group-level feature ${F}^{k}_{l}$ such that ${Q}^{k}_{l}$, where each elements in ${Q}^{k}_{l}=[q_{1}^{k}, ... ,q_{cls}^{k}]$ indicates the probability distribution of containing a particular categories at $k^{th}$ group .
We then encourage a semantic consistency among class distribution by utilizing l2-norm $\begin{Vmatrix}\cdot\end{Vmatrix}_{2}$ as follows: 
\begin{equation}\label{equ:1}
  \mathcal{L}_{co}(G,C) = \sum_{k \in \{1,...,K\} } \;\begin{Vmatrix} {Q}_{S}^{k}-{Q}_{T}^{k}\end{Vmatrix}^{2}.
\end{equation}

Minimizing loss \equref{equ:1} has two desirable effects.
First, it encourages the difference of each class distribution of group to be similar, and it also provides the supervisory signals for aligning the probability distribution of group-level features.

\begin{figure*}
   \centering
   \includegraphics[width=1 \linewidth]{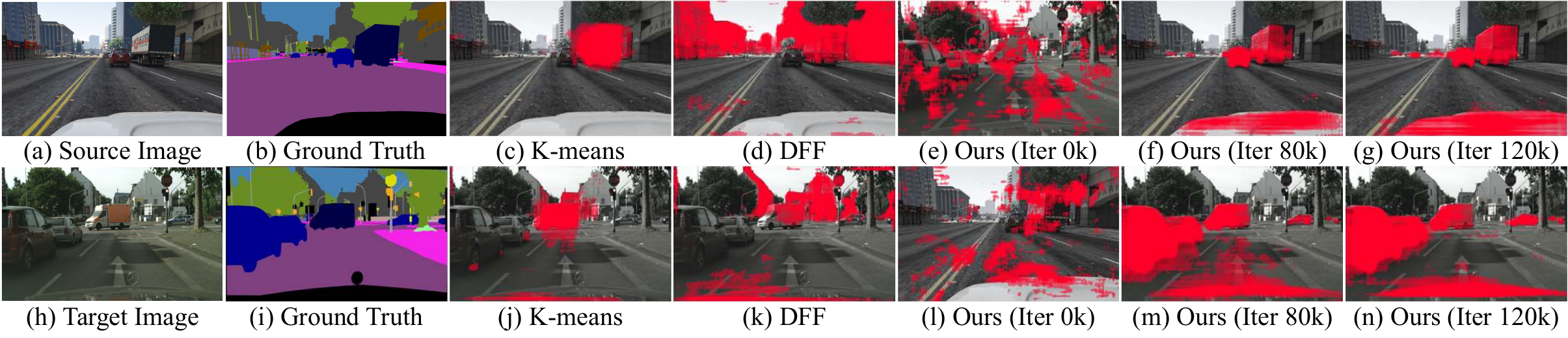}
   \caption{Visualization of cross-domain grouping result of (a) source and (c) target image with GT classes as corresponding colors (b,i)
   via (c,j) K-means, (d,f) DFF, (e,l) Ours (Iter 0k), (f,m) Ours (Iter 80k) and (g,n) Ours (Iter 120k).
    Compared to non-learnable model, our method can better capture semantic consistent objects across source and target domains.
    }
 
\label{fig:k_means vs dff vs ours}
\end{figure*}

\subsubsection{Orthogonality.}\label{sec:332}
The semantic consistency constraint in \equref{equ:1} encourages the class distribution of group across the source and target domains to be consistent.
This, however, does not guarantee that class distribution is different for each group.
In other words, we cannot divide the multi-modal complex distribution into several simple distributions.
To this end, we draw the second insight by introducing orthogonality constraint such that, any two class distribution ${Q}^{j_{1}}_{l}$ and ${Q}^{j_{2}}_{l}$, should be orthogonal each other.
It can be realized that their cosine similarity \equref{equ:cos} is 0 since ${Q}^{k}_{l}$ are non-negative value. 
We define the cosine similarity with l2-norm $\begin{Vmatrix}\cdot\end{Vmatrix}_{2}$ as follows:
\begin{equation}\label{equ:cos}
  \cos(Q_{l}^{j_{1}},Q_{l}^{j_{2}}) = {Q_{l}^{j_{1}}  \cdot Q_{l}^{j_{2}} \over \begin{Vmatrix} Q_{l}^{j_{1}} \end{Vmatrix}_{2}\begin{Vmatrix}  Q_{l}^{j_{2}} \end{Vmatrix}_{2}}, \; \; \;  j_{1},j_{2} \in \{  1, ..., K \}.
\end{equation}
We then formulate an orthogonal loss for training such that
\begin{equation}
  \mathcal{L}_{orth}(G,C) = \sum_{l} \sum_{{j}_{1},{j}_{2}} \; \cos({Q}_{l}^{j_{1}},{Q}_{l}^{j_{2}}),
\end{equation}
where we apply a loss function on each domain $l\in\left\{S,T\right\}$.
By forcing the cross-domain grouping module $C$ to make each group to be orthogonal, it can divide a multi-modal complex distribution into the $K$ simple class distributions.
\subsection{Losses for Cross-Domain Alignment}\label{sec:34}
In this section, we present a group-level adversarial learning framework as an alternative to \emph{global} domain adaptation and \emph{category-level} domain adaptation.


\subsubsection{Group-level alignment.}\label{sec:341}
To achieve group-level domain alignment, a straight forward method is to use $K$ independent discriminators, similar to conventional category-level domain alignment methods~\cite{Yi17,Liang19}.
However, we simultaneously update grouping module $C$
while training the overall network, thus cluster assignment may not be consistent at each training iteration.
To this end, we adopt conditional adversarial learning framework following ~\cite{Mingsheng18}, by combining group-level feature ${F}^{k}_{l}$ with ${Q}^{k}_{l}$ as a condition as follows:
\begin{align}\label{equ:overall}
\begin{split}
    \mathcal{L}_{cadv}(G,C,D) =-\sum_{k} \;[\log(D(F_{S}^{k} \otimes Q^{k}_{S}))] \\ \;\; -\sum_{k} \;[\log(1-D(F_{T}^{k}\otimes Q^{k}_{T}))],
    \end{split}
\end{align}
where ${\otimes}$ represent outer product operation.
Note that using group-level feature $F^{k}_{l}$ only as input to the discriminator is equivalent to global alignment, while we give a condition by using cross-covariance between $F^{k}_{l}$ and $Q^{K}_{l}$ as input.
This leads to discriminative domain alignment according to the different groups.

\subsubsection{Group-level class equivalence.}\label{sec:342}
For group-level adversarial learning, the existence of particular classes across different domains is desirable.
However, since the number of pixels for particular classes are dominant in each image, it can cause class imbalance problem.
Thus adaptation model tends to be biased towards majority classes and ignore minority classes~\cite{Yang18}.
To alleviate this, we propose group level class equivalence following Zhao et al.~\shortcite{Xiangyun18}.
We first apply max pooling layer for each group level feature $M_{l}^{k}$ such that each element of $M_{l}^{k} = [m_{l,1}^{k}, ... ,m_{l,cls}^{k}]$ is a maximum score for each category corresponding to group $k$.
We then utilize maximum classification score in the source domain ${m}_{S}^{k}$ as a pseudo-label, where we aim to train maximum classification score in target domain ${m}_{T}^{k}$ to be similar.
To this end, we apply multi-class binary cross-entropy loss for each class as follows :
\begin{equation}\label{equ:equivalence}
	\begin{split}
  \mathcal{L}_{cl}(G,C) = - \sum_{k} \; \sum_{u} [m\,_{S,u}^{k} \geq \tau] \; \log(m\,_{T,u}^{k}),
    \end{split}
\end{equation}
where Iverson bracket indicator function $[$·$]$ evaluates to 1 when it is true and 0 otherwise, and $u\in\{1,...,cls\}$ denotes category.
Note that we merely exclude too low probability with the threshold parameter $\tau$.

\subsection{Training}\label{sec:35}
The overall loss function of our approach can be written as 
\begin{flalign}\label{equ:overall}
	\begin{split}
 L(G,C,D)&=L_{seg}(G)
 +\lambda_{co}L_{co}(G,C)+\lambda_{cl}L_{cl}(G,C)\\ &+\lambda_{orth}L_{orth}(G,C) +\lambda_{cadv}L_{cadv}(G,C,D),
    \end{split}
\end{flalign}
where $L_{seg}$ is the supervised cross-entropy loss for semantic segmentation network on the source data, and $\lambda_{co},\lambda_{orth},\lambda_{cadv}$ and $\lambda_{cl}$ are balancing parameter for different losses.
We then solve the following minmax problem for optimizing $G,C$ and $D$.

\begin{equation}\label{equ:minmax}
	\begin{split}
 \min_{G,C} \max_{D}L(G,C,D).
    \end{split}
\end{equation}

\section{Experiments}\label{sec:4}

\subsection{Experimental Setting}\label{sec:41}
\subsubsection{Implementation details.}
The proposed method was implemented in PyTorch library~\cite{Adam17} and simulated on a PC with a single RTX Titan GPU.
We utilize BDL~\cite{Yunsheng19} as our baseline model following conventional work~\cite{Zhonghao20}, including self-supervised learning and image transferring framework.
To train the segmentation network, we utilize stochastic gradient descent (SGD) ~\shortcite{lecun1998gradient}, where the learning rate is set to $2.5 \times {10}^{-4}$.
For grouping network, we utilize SGD, with learning rate as $1 \times {10}^{-3}$.
Both learning rates decreased with ``poly'' learning rate policy with power fixed to 0.9 and momentum as 0.9.
For discriminator training, we use Adam~\shortcite{Diederik14} optimizer with an initial learning rate $1 \times {10}^{-4}$.
We jointly train our segmentation network, grouping network, and discriminator using \equref{equ:minmax} for a total of $120k$ iterations.
We randomly paired source and target images in each iteration.
Through the cross-validation using grid-search in log-scale, we set the hyper-parameters $\lambda _{co},\lambda _{orth},\lambda _{cadv},\lambda _{cl}$ and $\tau$ as 0.001, 0.001, 0.001, 0.0001 and 0.05, respectively.

\subsubsection{Datasets.}
For experiments, we use the GTA5~\cite{Stephan16} and SYNTHIA~\cite{German16} as source dataset.
GTA5 dataset~\cite{Stephan16} contains 24,966 images with 1914$\times$1052 resolution.
We resize images to 1280 $\times$ 760 following other work~\cite{Yi18}.
For SYNTHIA~\cite{German16}, we use SYNTHIA-RAND-CITYSCAPES dataset with 9,400 images with 1280$\times$760 resolution.
We use Cityscapes~\cite{Marius16} as target dataset, which consists of 2,975, 500 and 1,525 images with training, validation and test set.
We train our network with training set, while evaluation is done using validation set.
We resize images to 1024 $\times$ 512 for both training and testing as ~\cite{Yunsheng19}.
We evaluate the class-level intersection over union (IoU) and mean IoU (mIoU)~\cite{Mark15}.

\subsection{Analysis}\label{sec:analysis}

We first visualize each group through cross-domain grouping in \figref{fig:NMF}.
Clustered groups for various $k$ showed that our networks clustered semantically consistent regions across the source and target domains through \equref{equ:1}.
Also, clustered regions along different $k$ indicate that our network effectively divided regions into different group using \equref{equ:cos}.

\begin{figure*}
   \centering
   \includegraphics[width=1 \linewidth]{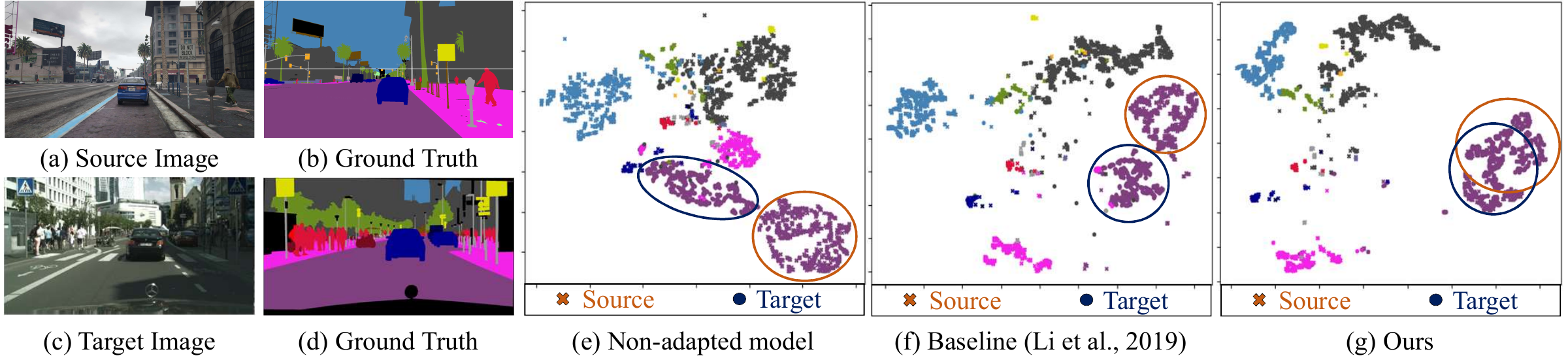}
   
   \caption{Visualization of output probability distribution of (a) source and (c) target image with GT classes as corresponding colors (b,d) via t-SNE using (e) non-adapted model, (f) baseline~\cite{Yunsheng19} and (g) ours. Our method effectively reduce domain discrepancy along with different domain, while others failed (represented using a circle).
   }
  
\label{fig:f8}
\end{figure*}

\begin{figure}
   \centering
   \includegraphics[width=0.99 \linewidth]{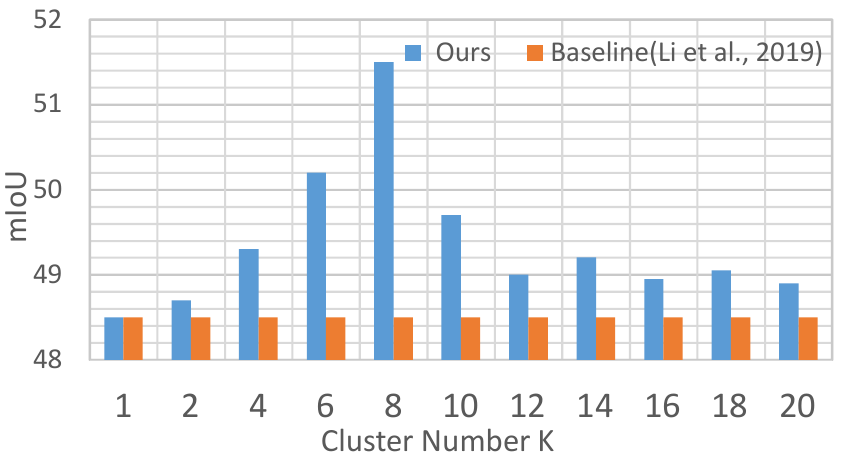}
     \caption{Ablation study for domain alignment with different number of clusters $K$ on GTA5 $\rightarrow$ Cityscapes.} 
   \label{fig:ablation_k}
\end{figure}

We further compare our grouping network with k-means clustering algorithm~\shortcite{Adam12} and deep feature factorization~\shortcite{Edo18} which is not trainable methods.
As shown in \figref{fig:k_means vs dff vs ours}, our method can better capture the object boundaries and semantic consistent objects across source and target domains compared to other non-learnable methods.
We further visualize each clustered group through cross-domain grouping with an evolving number of iteration.
As the number of iterations increases, cross-domain grouping and group-level domain alignment share complementary information, which decomposes the data distribution and aligns domains for each grouped sub-spaces in a joint and boosting manner.

In \figref{fig:f8}, we show the t-SNE visualization~\cite{Laurens08} of the output probability distribution of our method compares to the non-adapted method and baseline~\cite{Yunsheng19}.
The result shows that our method effectively aligns the distribution of the source and target domains, while others failed to reduce the domain discrepancy.
Furthermore, we observe that our model successfully grouped minority categories (i.e. traffic signs in yellow) while others failed.
It indicates that the loss \equref{equ:equivalence} can solve class imbalance problem.





\subsection{Ablation Study}\label{sec:ablation}
\figref{fig:ablation_k} shows the result of ablation experiments with different number of groups $K$.
Note that the results with $K=1$ are equivalent to global domain adaptation as a baseline.
The result shows that ours with various number of $K$ consistently outperform the baseline, which shows the effectiveness of our group-level domain alignment.
The performance has improved as $K$ increased from $1$, and after achieving the best performance at $K=8$ the rest showed no significant difference.
The lower performance with the larger number of $K$ indicates that over-clustered samples can actually degrade performance as conventional category-level adaptation methods.
Since the result with $K=8$ has shown the best performance on both GTA5 $\rightarrow$ Cityscapes and SYNTHIA $\rightarrow$ Cityscapes, we set $K$ as $8$ for all experiments.

\tabref{tab:ablation_loss} shows the result of ablation experiments to validate the effects of proposed loss functions.
It verifies the effectiveness of each loss function, including 
group-level domain adaptation, group-level semantic consistency, group-level orthogonality, and group-level class equivalence.
The full usage of our proposed loss functions yields the best results. 
We also find that adding group-level orthogonality leads to a large improvement in the performance, which demonstrates that we effectively divide the multi-modal complex distribution into $K$ simple distributions for group-level domain alignment.

\begin{table}[!t]
\centering
 \resizebox{0.48\textwidth}{!}{   
    \begin{tabular}{c|ccccc|c}
    \hline
    \multirow{2}{*}{Method} & \multicolumn{5}{c|}{Loss Functions} & \multirow{2}{*}{mIOU} \\ \cline{2-6}
     & ${L}_{seg}$ & ${L}_{cadv}$ & ${L}_{co}$ & ${L}_{orth}$ & ${L}_{cl}$ &\\ \hline \hline
    Source only & \checkmark & & & & &36.6 \\ \hline 
    \multirow{4}{*}{Ours} & \checkmark & \checkmark & & & &48.8\\
    & \checkmark & \checkmark & \checkmark & & &49.1 \\
    & \checkmark & \checkmark & \checkmark & \checkmark & &50.8\\
    & \checkmark & \checkmark & \checkmark & \checkmark & \checkmark &\textbf{51.5}\\\hline
    \end{tabular}
     }
        \caption{Ablation study for domain alignment with different loss functions on GTA5 $\rightarrow$ Cityscapes.
    }\label{tab:ablation_loss}
\end{table}

\begin{figure*}
   \centering
   \includegraphics[width=0.99 \linewidth]{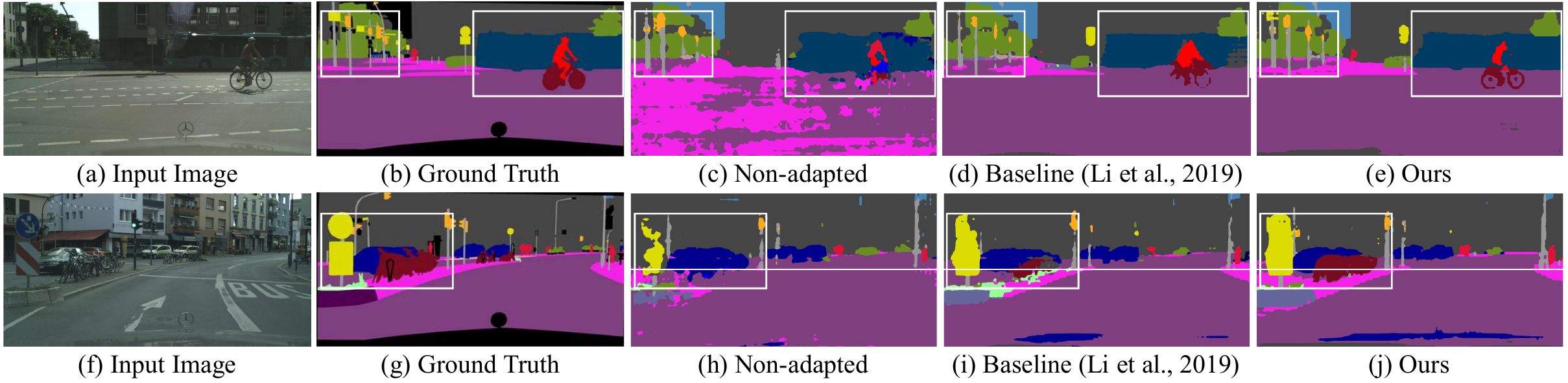}
   
     \caption{Qualitative results of domain adaptation on GTA5 $\rightarrow$ Cityscapes. 
    (From left to right) Input image, ground-truth, non-adapted result, baseline result and ours result.} 
   \label{fig:segmentation_output}
\end{figure*}

\begin{table*}[!t]
\centering
\makebox[1 \textwidth][c]{       
\resizebox{1 \textwidth}{!}{   
\begin{tabular}{c|ccccccccccccccccccc|c}
\hline
\multicolumn{21}{c}{\begin{scriptsize}GTA5 $\rightarrow$ Cityscapes\end{scriptsize}}\\ \hline
 &\begin{scriptsize}Road\end{scriptsize}
 &\begin{scriptsize}SW\end{scriptsize}
 &\begin{scriptsize}Build\end{scriptsize}
 &\begin{scriptsize}Wall\end{scriptsize}  
 &\begin{scriptsize}Fence\end{scriptsize} 
 &\begin{scriptsize}Pole\end{scriptsize}  
 &\begin{scriptsize}TL\end{scriptsize}
 &\begin{scriptsize}TS\end{scriptsize}  
 &\begin{scriptsize}Veg.\end{scriptsize}   
 &\begin{scriptsize}terrain\end{scriptsize} 
 &\begin{scriptsize}Sky\end{scriptsize} 
 &\begin{scriptsize}PR\end{scriptsize} 
 &\begin{scriptsize}Rider\end{scriptsize} 
 &\begin{scriptsize}Car\end{scriptsize}   
 &\begin{scriptsize}Truck\end{scriptsize} 
 &\begin{scriptsize}Bus\end{scriptsize}   
 &\begin{scriptsize}Train\end{scriptsize} 
 &\begin{scriptsize}Motor\end{scriptsize} 
 &\begin{scriptsize}Bike\end{scriptsize}  
 &\begin{scriptsize}mIoU\end{scriptsize}  \\ \hline \hline
\begin{scriptsize}Without Adaptation \end{scriptsize}
 & \begin{scriptsize}75.8\end{scriptsize} 
 & \begin{scriptsize}16.8\end{scriptsize}    
 & \begin{scriptsize}77.2\end{scriptsize}    
 & \begin{scriptsize}12.5\end{scriptsize} 
 & \begin{scriptsize}21.0\end{scriptsize}  
 &\begin{scriptsize}25.5\end{scriptsize}
 & \begin{scriptsize}30.1\end{scriptsize} 
 & \begin{scriptsize}20.1\end{scriptsize}
 & \begin{scriptsize}81.3\end{scriptsize}
 & \begin{scriptsize}24.6\end{scriptsize}   
 & \begin{scriptsize}70.3\end{scriptsize}
 & \begin{scriptsize}53.8\end{scriptsize} 
 & \begin{scriptsize}26.4\end{scriptsize} 
 & \begin{scriptsize}49.9\end{scriptsize} 
 & \begin{scriptsize}17.2\end{scriptsize}
 & \begin{scriptsize}25.9\end{scriptsize} 
 & \begin{scriptsize}6.5\end{scriptsize} 
 & \begin{scriptsize}25.3\end{scriptsize} 
 & \begin{scriptsize}36.0\end{scriptsize}
 & \begin{scriptsize}36.6\end{scriptsize} \\ \hline

\begin{scriptsize}Tsai et al.~\shortcite{Yi18} \end{scriptsize}
 & \begin{scriptsize}86.5\end{scriptsize}
 & \begin{scriptsize}36.0\end{scriptsize}    
 & \begin{scriptsize}79.9\end{scriptsize}    
 & \begin{scriptsize}23.4\end{scriptsize} 
 & \begin{scriptsize}23.3\end{scriptsize}  
 &\begin{scriptsize}23.9\end{scriptsize} 
 & \begin{scriptsize}35.2\end{scriptsize} 
 & \begin{scriptsize}14.8\end{scriptsize}
 & \begin{scriptsize}83.4\end{scriptsize} 
 & \begin{scriptsize}33.3\end{scriptsize}   
 & \begin{scriptsize}75.6\end{scriptsize} 
 & \begin{scriptsize}58.5\end{scriptsize} 
 & \begin{scriptsize}27.6\end{scriptsize} 
 & \begin{scriptsize}73.7\end{scriptsize} 
 & \begin{scriptsize}32.5\end{scriptsize} 
 & \begin{scriptsize}35.4\end{scriptsize} 
 & \begin{scriptsize}3.9\end{scriptsize}  
 & \begin{scriptsize}30.1\end{scriptsize}
 & \begin{scriptsize}28.1\end{scriptsize}
 & \begin{scriptsize}42.4\end{scriptsize} \\ \hline
  
\begin{scriptsize}Wu et al.~\shortcite{Zuxuan18} \end{scriptsize}
 & \begin{scriptsize}85.0\end{scriptsize}
 & \begin{scriptsize}30.8\end{scriptsize}   
 & \begin{scriptsize}81.3\end{scriptsize}     
 & \begin{scriptsize}25.8\end{scriptsize} 
 & \begin{scriptsize}21.2\end{scriptsize}  
 &\begin{scriptsize}22.2\end{scriptsize}
 & \begin{scriptsize}25.4\end{scriptsize} 
 & \begin{scriptsize}26.6\end{scriptsize}
 & \begin{scriptsize}83.4\end{scriptsize} 
 & \begin{scriptsize}36.7\end{scriptsize}  
 & \begin{scriptsize}76.2\end{scriptsize} 
 & \begin{scriptsize}58.9\end{scriptsize}  
 & \begin{scriptsize}24.9\end{scriptsize} 
 & \begin{scriptsize}80.7\end{scriptsize} 
 & \begin{scriptsize}29.5\end{scriptsize} 
 & \begin{scriptsize}42.9\end{scriptsize} 
 & \begin{scriptsize}2.5\end{scriptsize}  
 & \begin{scriptsize}26.9\end{scriptsize} 
 & \begin{scriptsize}11.6\end{scriptsize} 
 & \begin{scriptsize}41.7\end{scriptsize} \\ 
 
 \begin{scriptsize}Chang et al.~\shortcite{Chang19} \end{scriptsize}
 & \begin{scriptsize}91.5\end{scriptsize}
 & \begin{scriptsize}47.5\end{scriptsize}    
 & \begin{scriptsize}82.5\end{scriptsize}    
 & \begin{scriptsize}31.3\end{scriptsize} 
 & \begin{scriptsize}25.6\end{scriptsize}  
 &\begin{scriptsize}33.0\end{scriptsize} 
 & \begin{scriptsize}33.7\end{scriptsize} 
 & \begin{scriptsize}25.8\end{scriptsize}
 & \begin{scriptsize}82.7\end{scriptsize} 
 & \begin{scriptsize}28.8\end{scriptsize}   
 & \begin{scriptsize}82.7\end{scriptsize} 
 & \begin{scriptsize}62.4\end{scriptsize} 
 & \begin{scriptsize}30.8\end{scriptsize} 
 & \begin{scriptsize}85.2\end{scriptsize} 
 & \begin{scriptsize}27.7\end{scriptsize} 
 & \begin{scriptsize}34.5\end{scriptsize} 
 & \begin{scriptsize}6.4\end{scriptsize}  
 & \begin{scriptsize}25.2\end{scriptsize}
 & \begin{scriptsize}24.4\end{scriptsize} 
 & \begin{scriptsize}45.4\end{scriptsize} \\ 
 
 \begin{scriptsize}Li et al.~\shortcite{Yunsheng19} \end{scriptsize}
 & \begin{scriptsize}91.0\end{scriptsize}
 & \begin{scriptsize}44.7\end{scriptsize}    
 & \begin{scriptsize}84.2\end{scriptsize}    
 & \begin{scriptsize}\textbf{34.6}\end{scriptsize}
 & \begin{scriptsize}27.6\end{scriptsize}
 &\begin{scriptsize}30.2\end{scriptsize} 
 & \begin{scriptsize}36.0\end{scriptsize} 
 & \begin{scriptsize}36.0\end{scriptsize}
 & \begin{scriptsize}\textbf{85.0}\end{scriptsize} 
 & \begin{scriptsize}\textbf{43.6}\end{scriptsize}   
 & \begin{scriptsize}83.0\end{scriptsize} 
 & \begin{scriptsize}58.6\end{scriptsize}
 & \begin{scriptsize}31.6\end{scriptsize} 
 & \begin{scriptsize}83.3\end{scriptsize} 
 & \begin{scriptsize}35.3\end{scriptsize} 
 & \begin{scriptsize}49.7\end{scriptsize} 
 & \begin{scriptsize}3.3\end{scriptsize} 
 & \begin{scriptsize}28.8\end{scriptsize}
 & \begin{scriptsize}35.6\end{scriptsize}
 & \begin{scriptsize}48.5\end{scriptsize} \\ \hline

\begin{scriptsize}Luo et al.~\shortcite{Yawei19} \end{scriptsize}
 & \begin{scriptsize}87.0\end{scriptsize}
 & \begin{scriptsize}27.1\end{scriptsize}    
 & \begin{scriptsize}79.6\end{scriptsize}    
 & \begin{scriptsize}27.3\end{scriptsize} 
 & \begin{scriptsize}23.3\end{scriptsize}  
 &\begin{scriptsize}28.3\end{scriptsize} 
 & \begin{scriptsize}35.5\end{scriptsize} 
 & \begin{scriptsize}24.2\end{scriptsize}
 & \begin{scriptsize}83.6\end{scriptsize} 
 & \begin{scriptsize}27.4\end{scriptsize}   
 & \begin{scriptsize}74.2\end{scriptsize} 
 & \begin{scriptsize}58.6\end{scriptsize} 
 & \begin{scriptsize}28.0\end{scriptsize} 
 & \begin{scriptsize}76.2\end{scriptsize} 
 & \begin{scriptsize}33.1\end{scriptsize} 
 & \begin{scriptsize}36.7\end{scriptsize} 
 & \begin{scriptsize}\textbf{6.7}\end{scriptsize}  
 & \begin{scriptsize}31.9\end{scriptsize}
 & \begin{scriptsize}31.4\end{scriptsize} 
 & \begin{scriptsize}43.2\end{scriptsize} \\
 
 \begin{scriptsize}Du et al.~\shortcite{Liang19} \end{scriptsize}
 & \begin{scriptsize}90.3\end{scriptsize}
 & \begin{scriptsize}38.9\end{scriptsize}    
 & \begin{scriptsize}81.7\end{scriptsize}    
 & \begin{scriptsize}24.8\end{scriptsize} 
 & \begin{scriptsize}22.9\end{scriptsize}  
 &\begin{scriptsize}30.5\end{scriptsize}
 & \begin{scriptsize}37.0\end{scriptsize} 
 & \begin{scriptsize}21.2\end{scriptsize}
 & \begin{scriptsize}84.8\end{scriptsize} 
 & \begin{scriptsize}38.8\end{scriptsize}   
 & \begin{scriptsize}76.9\end{scriptsize} 
 & \begin{scriptsize}58.8\end{scriptsize} 
 & \begin{scriptsize}30.7\end{scriptsize} 
 & \begin{scriptsize}85.7\end{scriptsize}
 & \begin{scriptsize}30.6\end{scriptsize} 
 & \begin{scriptsize}38.1\end{scriptsize} 
 & \begin{scriptsize}5.9\end{scriptsize}  
 & \begin{scriptsize}28.3\end{scriptsize}
 & \begin{scriptsize}36.9\end{scriptsize}
 & \begin{scriptsize}45.4\end{scriptsize} \\
 
\begin{scriptsize}Vu et al.~\shortcite{Tuan19} \end{scriptsize}
 & \begin{scriptsize}90.3\end{scriptsize}
 & \begin{scriptsize}38.9\end{scriptsize}    
 & \begin{scriptsize}81.7\end{scriptsize}    
 & \begin{scriptsize}24.8\end{scriptsize} 
 & \begin{scriptsize}22.9\end{scriptsize}  
 &\begin{scriptsize}30.5\end{scriptsize}
 & \begin{scriptsize}37.0\end{scriptsize} 
 & \begin{scriptsize}21.2\end{scriptsize}
 & \begin{scriptsize}84.8\end{scriptsize} 
 & \begin{scriptsize}38.8\end{scriptsize}   
 & \begin{scriptsize}76.9\end{scriptsize} 
 & \begin{scriptsize}58.8\end{scriptsize} 
 & \begin{scriptsize}30.7\end{scriptsize} 
 & \begin{scriptsize}85.7\end{scriptsize}
 & \begin{scriptsize}30.6\end{scriptsize} 
 & \begin{scriptsize}38.1\end{scriptsize} 
 & \begin{scriptsize}5.9\end{scriptsize}  
 & \begin{scriptsize}28.3\end{scriptsize}
 & \begin{scriptsize}36.9\end{scriptsize}
 &  \begin{scriptsize}45.4\end{scriptsize} \\ 
 
\begin{scriptsize}Tsai et al.~\shortcite{Yi19} \end{scriptsize}
 & \begin{scriptsize}92.3\end{scriptsize} 
 & \begin{scriptsize}51.9\end{scriptsize}
 & \begin{scriptsize}82.1\end{scriptsize}    
 & \begin{scriptsize}29.2\end{scriptsize} 
 & \begin{scriptsize}25.1\end{scriptsize}  
 &\begin{scriptsize}24.5\end{scriptsize} 
 & \begin{scriptsize}33.8\end{scriptsize} 
 & \begin{scriptsize}33.0\end{scriptsize}
 & \begin{scriptsize}82.4\end{scriptsize} 
 & \begin{scriptsize}32.8\end{scriptsize}   
 & \begin{scriptsize}82.2\end{scriptsize} 
 & \begin{scriptsize}58.6\end{scriptsize} 
 & \begin{scriptsize}27.2\end{scriptsize} 
 & \begin{scriptsize}84.3\end{scriptsize} 
 & \begin{scriptsize}33.4\end{scriptsize} 
 & \begin{scriptsize}46.3\end{scriptsize} 
 & \begin{scriptsize}2.2\end{scriptsize}
 & \begin{scriptsize}29.5\end{scriptsize}
 & \begin{scriptsize}32.3\end{scriptsize}
 & \begin{scriptsize}46.5\end{scriptsize} \\  
 
 \begin{scriptsize}Huang et al.~\shortcite{Jiaxing20} \end{scriptsize}
 & \begin{scriptsize}\textbf{92.4}\end{scriptsize} 
 & \begin{scriptsize}\textbf{55.3}\end{scriptsize}
 & \begin{scriptsize}82.3\end{scriptsize}    
 & \begin{scriptsize}31.2\end{scriptsize} 
 & \begin{scriptsize}\textbf{29.1}\end{scriptsize}  
 &\begin{scriptsize}32.5\end{scriptsize} 
 & \begin{scriptsize}33.2\end{scriptsize} 
 & \begin{scriptsize}35.6\end{scriptsize}
 & \begin{scriptsize}83.5\end{scriptsize} 
 & \begin{scriptsize}34.8\end{scriptsize}   
 & \begin{scriptsize}84.2\end{scriptsize} 
 & \begin{scriptsize}58.9\end{scriptsize} 
 & \begin{scriptsize}32.2\end{scriptsize} 
 & \begin{scriptsize}84.7\end{scriptsize} 
 & \begin{scriptsize}40.6\end{scriptsize} 
 & \begin{scriptsize}46.1\end{scriptsize} 
 & \begin{scriptsize}2.1\end{scriptsize}
 & \begin{scriptsize}31.1\end{scriptsize}
 & \begin{scriptsize}32.7\end{scriptsize}
 & \begin{scriptsize}48.6\end{scriptsize} \\  
 
\begin{scriptsize}Wang et al.~\shortcite{Zhonghao20} \end{scriptsize}
 & \begin{scriptsize}90.6\end{scriptsize} 
 & \begin{scriptsize}44.7\end{scriptsize}
 & \begin{scriptsize}\textbf{84.8}\end{scriptsize}    
 & \begin{scriptsize}34.3\end{scriptsize} 
 & \begin{scriptsize}28.7\end{scriptsize}  
 &\begin{scriptsize}31.6\end{scriptsize} 
 & \begin{scriptsize}35.0\end{scriptsize} 
 & \begin{scriptsize}37.6\end{scriptsize}
 & \begin{scriptsize}84.7\end{scriptsize} 
 & \begin{scriptsize}43.3\end{scriptsize}   
 & \begin{scriptsize}\textbf{85.3}\end{scriptsize} 
 & \begin{scriptsize}57.0\end{scriptsize} 
 & \begin{scriptsize}31.5\end{scriptsize} 
 & \begin{scriptsize}83.8\end{scriptsize} 
 & \begin{scriptsize}42.6\end{scriptsize} 
 & \begin{scriptsize}48.5\end{scriptsize} 
 & \begin{scriptsize}1.9\end{scriptsize}
 & \begin{scriptsize}30.4\end{scriptsize}
 & \begin{scriptsize}39.0\end{scriptsize}
 & \begin{scriptsize}49.2\end{scriptsize} \\  \hline

\begin{scriptsize} Ours \end{scriptsize}            
& \begin{scriptsize} 91.1 \end{scriptsize}
&  \begin{scriptsize}52.8 \end{scriptsize}   
& \begin{scriptsize}84.6\end{scriptsize}
&  \begin{scriptsize} 32.0\end{scriptsize}
&  \begin{scriptsize} 27.1\end{scriptsize}
&  \begin{scriptsize} \textbf{33.8}\end{scriptsize} 
&  \begin{scriptsize} \textbf{38.4}\end{scriptsize}
&  \begin{scriptsize} \textbf{40.3}\end{scriptsize} 
&  \begin{scriptsize} 84.6\end{scriptsize} 
&  \begin{scriptsize} 42.8\end{scriptsize}   
&  \begin{scriptsize} 85.0\end{scriptsize} 
&  \begin{scriptsize} \textbf{64.2}\end{scriptsize}  
&  \begin{scriptsize} \textbf{36.5}\end{scriptsize}
&  \begin{scriptsize} \textbf{87.3}\end{scriptsize}
& \begin{scriptsize}\textbf{44.4}\end{scriptsize}
& \begin{scriptsize} \textbf{51.0}\end{scriptsize} 
&  \begin{scriptsize} 0.0\end{scriptsize} 
& \begin{scriptsize} \textbf{37.3}\end{scriptsize} 
&  \begin{scriptsize} \textbf{44.9}\end{scriptsize} 
&  \begin{scriptsize} \textbf{51.5}\end{scriptsize} \\ \hline
\end{tabular}}}

\caption{Quantitative results of domain adaptation on GTA5 $\rightarrow$ Cityscapes.}
\label{tab:gta5_cityscape}
\end{table*}

\begin{table*}[!t]
\makebox[1 \textwidth][c]{       
\resizebox{1 \textwidth}{!}{   
\begin{tabular}{c|ccccccccccccc|c}
\hline
\multicolumn{15}{c}{\begin{scriptsize}SYNTHIA $\rightarrow$  Cityscapes\end{scriptsize}} \\\hline              
 & \begin{scriptsize}Road\end{scriptsize}
 &\begin{scriptsize}SW\end{scriptsize}
 &\begin{scriptsize}Build\end{scriptsize}
 &\begin{scriptsize}TL\end{scriptsize} 
 &\begin{scriptsize}TS\end{scriptsize}  
 &\begin{scriptsize}Veg.\end{scriptsize}   
 &\begin{scriptsize}Sky\end{scriptsize}   
 &\begin{scriptsize}PR\end{scriptsize} 
 &\begin{scriptsize}Rider\end{scriptsize} 
 &\begin{scriptsize}Car\end{scriptsize}   
 &\begin{scriptsize}Bus\end{scriptsize}   
 &\begin{scriptsize}Motor\end{scriptsize} 
 &\begin{scriptsize}Bike\end{scriptsize}  
 &\begin{scriptsize}mIoU\end{scriptsize}  \\ \hline \hline

\begin{scriptsize}Tsai et al.~\shortcite{Yi18}\end{scriptsize}
 & \begin{scriptsize}84.3\end{scriptsize} 
 & \begin{scriptsize}42.7\end{scriptsize}    
 & \begin{scriptsize}77.5\end{scriptsize}    
 & \begin{scriptsize}4.7\end{scriptsize}
 & \begin{scriptsize}7.0\end{scriptsize} 
 &\begin{scriptsize}77.9\end{scriptsize} 
 & \begin{scriptsize}82.5\end{scriptsize} 
 & \begin{scriptsize}54.3\end{scriptsize}
 & \begin{scriptsize}21.0\end{scriptsize}
 & \begin{scriptsize}72.3\end{scriptsize}   
 & \begin{scriptsize}32.2\end{scriptsize}
 & \begin{scriptsize}18.9\end{scriptsize} 
 & \begin{scriptsize}32.3\end{scriptsize} 
 & \begin{scriptsize}46.7\end{scriptsize} \\ \hline
 
 \begin{scriptsize}Li et al.~\shortcite{Yunsheng19}\end{scriptsize}
 & \begin{scriptsize}86.0\end{scriptsize}
 & \begin{scriptsize}46.7\end{scriptsize}    
 & \begin{scriptsize}80.3\end{scriptsize}    
 & \begin{scriptsize}14.1\end{scriptsize} 
 & \begin{scriptsize}11.6\end{scriptsize}  
 &\begin{scriptsize}79.2\end{scriptsize} 
 & \begin{scriptsize}81.3\end{scriptsize} 
 & \begin{scriptsize}54.1\end{scriptsize}
 & \begin{scriptsize}27.9\end{scriptsize}
 & \begin{scriptsize}73.7\end{scriptsize}   
 & \begin{scriptsize}42.2\end{scriptsize}
 & \begin{scriptsize}25.7\end{scriptsize} 
 & \begin{scriptsize}45.3\end{scriptsize} 
 & \begin{scriptsize}51.4\end{scriptsize} \\ \hline

 \begin{scriptsize}Luo et al.~\shortcite{Yawei19} \end{scriptsize}
 & \begin{scriptsize}82.5\end{scriptsize} 
 & \begin{scriptsize}24.0\end{scriptsize}    
 & \begin{scriptsize}79.4\end{scriptsize}    
 & \begin{scriptsize}16.5\end{scriptsize} 
 & \begin{scriptsize}12.7\end{scriptsize}  
 &\begin{scriptsize}79.2\end{scriptsize} 
 & \begin{scriptsize}82.8\end{scriptsize} 
 & \begin{scriptsize}58.3\end{scriptsize}
 & \begin{scriptsize}18.0\end{scriptsize}
 & \begin{scriptsize}79.3\end{scriptsize}   
 & \begin{scriptsize}25.3\end{scriptsize}
 & \begin{scriptsize}17.6\end{scriptsize} 
 & \begin{scriptsize}25.9\end{scriptsize} 
  &\begin{scriptsize}46.3\end{scriptsize} \\

  \begin{scriptsize}Du et al.~\shortcite{Liang19}\end{scriptsize}
 & \begin{scriptsize}84.6\end{scriptsize} 
 & \begin{scriptsize}41.7\end{scriptsize}    
 & \begin{scriptsize}80.8\end{scriptsize}
 & \begin{scriptsize}11.5\end{scriptsize}
 & \begin{scriptsize}14.7\end{scriptsize}  
 &\begin{scriptsize}80.8\end{scriptsize}
 & \begin{scriptsize}85.3\end{scriptsize} 
 & \begin{scriptsize}57.5\end{scriptsize}
 & \begin{scriptsize}21.6\end{scriptsize}
 & \begin{scriptsize}\textbf{82.0}\end{scriptsize}
 & \begin{scriptsize}36.0\end{scriptsize}
 & \begin{scriptsize}19.3\end{scriptsize} 
 & \begin{scriptsize}34.5\end{scriptsize} 
 & \begin{scriptsize}50.0\end{scriptsize} \\ 
 
  \begin{scriptsize} Huang et al.~\shortcite{Jiaxing20}\end{scriptsize}
 & \begin{scriptsize}86.2\end{scriptsize}
 & \begin{scriptsize}44.9\end{scriptsize}    
 & \begin{scriptsize}79.5\end{scriptsize}    
 & \begin{scriptsize}9.4\end{scriptsize} 
 & \begin{scriptsize}11.8\end{scriptsize}  
 &\begin{scriptsize}78.6\end{scriptsize} 
 & \begin{scriptsize}\textbf{86.5}\end{scriptsize} 
 & \begin{scriptsize}57.2\end{scriptsize}
 & \begin{scriptsize}26.1\end{scriptsize}
 & \begin{scriptsize}76.8\end{scriptsize}   
 & \begin{scriptsize}39.9\end{scriptsize}
 & \begin{scriptsize}21.5\end{scriptsize} 
 & \begin{scriptsize}32.1\end{scriptsize} 
 & \begin{scriptsize}50.0\end{scriptsize}\\ 

 \begin{scriptsize} Wang et al.~\shortcite{Zhonghao20}\end{scriptsize}
 & \begin{scriptsize}83.0\end{scriptsize}
 & \begin{scriptsize}44.0\end{scriptsize}    
 & \begin{scriptsize}80.3\end{scriptsize}    
 & \begin{scriptsize}17.1\end{scriptsize} 
 & \begin{scriptsize}15.8\end{scriptsize}  
 &\begin{scriptsize}80.5\end{scriptsize} 
 & \begin{scriptsize}81.8\end{scriptsize} 
 & \begin{scriptsize}\textbf{59.9}\end{scriptsize}
 & \begin{scriptsize}33.1\end{scriptsize}
 & \begin{scriptsize}70.2\end{scriptsize}   
 & \begin{scriptsize}37.3\end{scriptsize}
 & \begin{scriptsize}28.5\end{scriptsize} 
 & \begin{scriptsize}45.8\end{scriptsize} 
 & \begin{scriptsize}52.1\end{scriptsize}\\ \hline

\begin{scriptsize} Ours \end{scriptsize}
& \begin{scriptsize} \textbf{90.7} \end{scriptsize} 
&  \begin{scriptsize} \textbf{49.5} \end{scriptsize}   
& \begin{scriptsize} \textbf{84.5}\end{scriptsize}   
&  \begin{scriptsize} \textbf{33.6} \end{scriptsize} 
&  \begin{scriptsize} \textbf{38.9}\end{scriptsize} 
&  \begin{scriptsize} \textbf{84.6}\end{scriptsize} 
&  \begin{scriptsize} 84.6\end{scriptsize} 
&  \begin{scriptsize}  59.8\end{scriptsize}  
&  \begin{scriptsize} \textbf{33.3}\end{scriptsize}
& \begin{scriptsize} 80.8\end{scriptsize}
& \begin{scriptsize} \textbf{51.5}\end{scriptsize} 
& \begin{scriptsize} \textbf{37.6}\end{scriptsize} 
&  \begin{scriptsize}\textbf{45.9}\end{scriptsize} 
&  \begin{scriptsize} \textbf{54.1}\end{scriptsize}\\ \hline
\end{tabular}}}

\caption{Quantitative results of domain adaptation on SYNTHIA $\rightarrow$ Cityscapes.}
\label{tab:synthia_cityscape}
\end{table*}

\subsection{Comparison with state-of-the-art methods}

\subsubsection{GTA5 $\rightarrow$ Cityscapes.}
In the following, we evaluated our method on GTA5 $\rightarrow$ Cityscapes in comparison to the state-of-the-art methods including without adaptation,  global adaptation ~\cite{Yi18}, image-level adaptation~\cite{Zuxuan18,Chang19,Yunsheng19} and category-level domain alignment~\cite{Yawei19,Liang19,Tuan19,Yi19,Jiaxing20,Zhonghao20}.
As shown in \tabref{tab:gta5_cityscape}, our method outperforms all other models on the categories ``car, truck, bus, motor, and bike'' which share similar appearance.
Specifically, we observe that our model achieves performance improvement on the categories ``pole and traffic sign''.
It demonstrates that our group-level class equivalence effectively solves the class imbalance problem.

\subsubsection{SYNTHIA $\rightarrow$ Cityscapes.}
We further compared our method to the state-of-the-art methods \cite{Yawei19,Yi18,Liang19,Yunsheng19,Jiaxing20,Zhonghao20} on SYNTHIA $\rightarrow$ Cityscapes, where 13 common classes between SYNTHIA \cite{German16} and Cityscapes \cite{Marius16} datasets are evaluated.
As shown in \tabref{tab:synthia_cityscape}, our method outperforms conventional methods.
We can observe that our model achieve similar improvements in GTA5 $\rightarrow$ Cityscapes scenario.
Compared to baseline~\cite{Yunsheng19}, our model achieves large improvement in the performance ``traffic sign and traffic light''.

\section{Conclusion}
We have introduced cross-domain grouping and alignment for domain adaptive semantic segmentation.
The key idea is to apply an intermediate grouping module such that multi-modal data distribution can be divided into several simple distributions.
We then apply group-level domain alignment across source and target domains, where the grouping network and segmentation network can be trained in a joint and boosting manner using semantic consistency and orthogonality constraints.
To solve the class imbalance problem, we have further introduced a group-level class equivalence constraint, resulting state-of-the-art performance on domain adaptive semantic segmentation.
We believe our approach will facilitate further advances in unsupervised domain adaptation on various computer vision tasks.

 \subsubsection{Acknowledgements.}
 This research was supported by R\&D program for Advanced Integrated-intelligence for Identification (AIID) through the National Research Foundation of KOREA (NRF) funded by Ministry of Science and ICT (NRF-2018M3E3A1057289).

\bibliographystyle{aaai21}
\bibliography{egbib}

\end{document}


\maketitle
In this supplementary material, we provide additional results including visualization of cross-domain grouping with evolving iterations, with random pairs, visualization of class distribution for each group, and more qualitative results.
\subsubsection{Visualization of clustered groups with evolving iterations}\label{sec:1}
\begin{figure*}
   \centering
   \includegraphics[width=0.99 \linewidth]{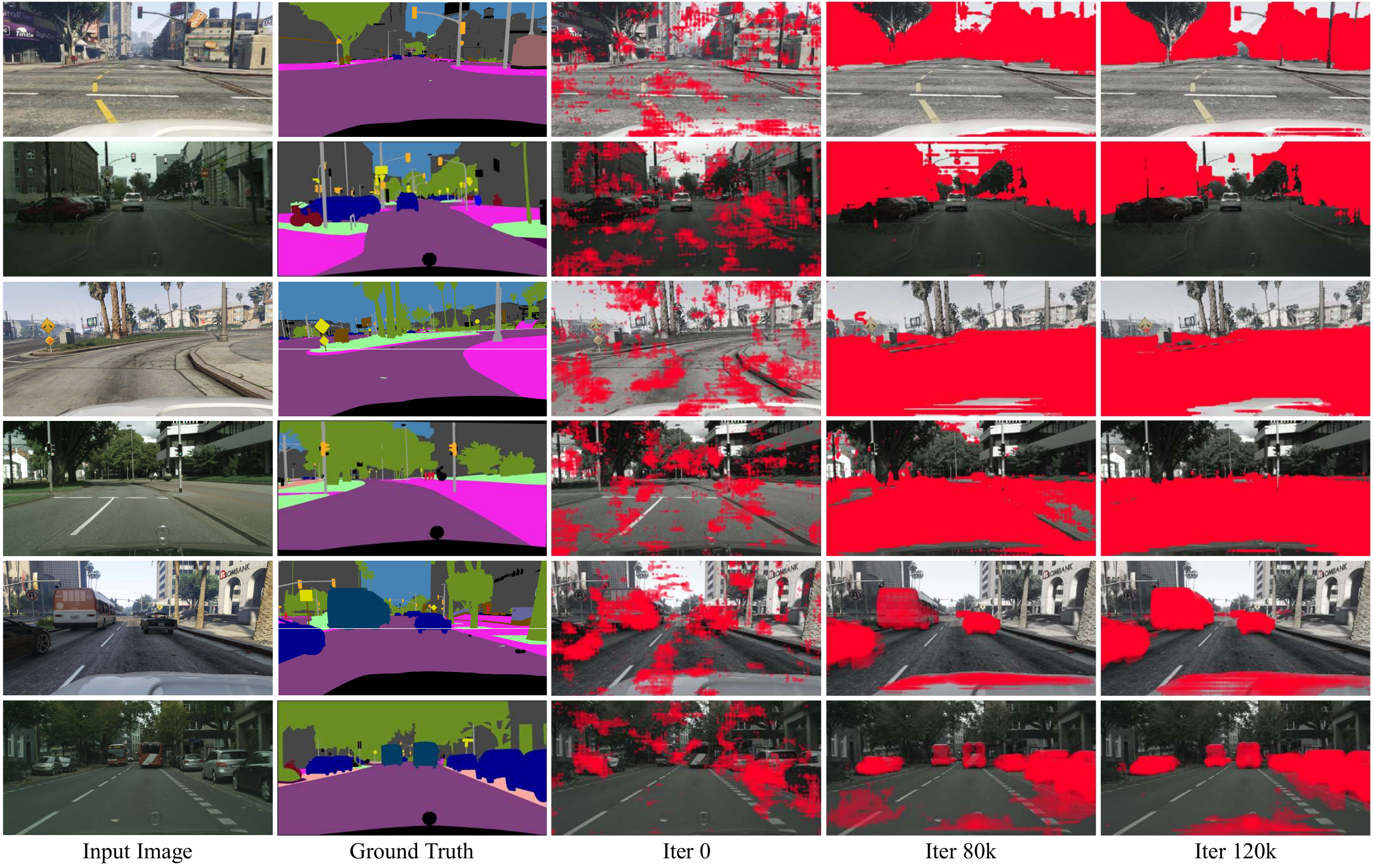}
   
     \caption{Visualization of clustered groups on source (1$^{st}$, 3$^{rd}$, 5$^{th}$) and target (2$^{nd}$, 4$^{th}$, 6$^{th}$) image with evolving iterations.
     }
   \label{fig:grouping_iter}
\end{figure*}

	In \figref{fig:grouping_iter}, we visualize each clustered group through cross-domain grouping with an evolving number of the iterations.
	As the number of iteration increases, our cross-domain grouping module can better capture object boundaries and semantic consistent objects across the source and target domains.
	We can observe that categories with small inter-class variation (e.g. car and bus, road, and sidewalk) are clustered together through our grouping module.
	This demonstrates that the cross-domain grouping module can mitigate inaccurate predictions by clustering samples with inter-class ambiguities for group-level domain adaptation.
	

\subsubsection{Visualization of clustered groups with random pairs}\label{sec:2}
\begin{figure*}
   \centering
   \includegraphics[width=0.99 \linewidth]{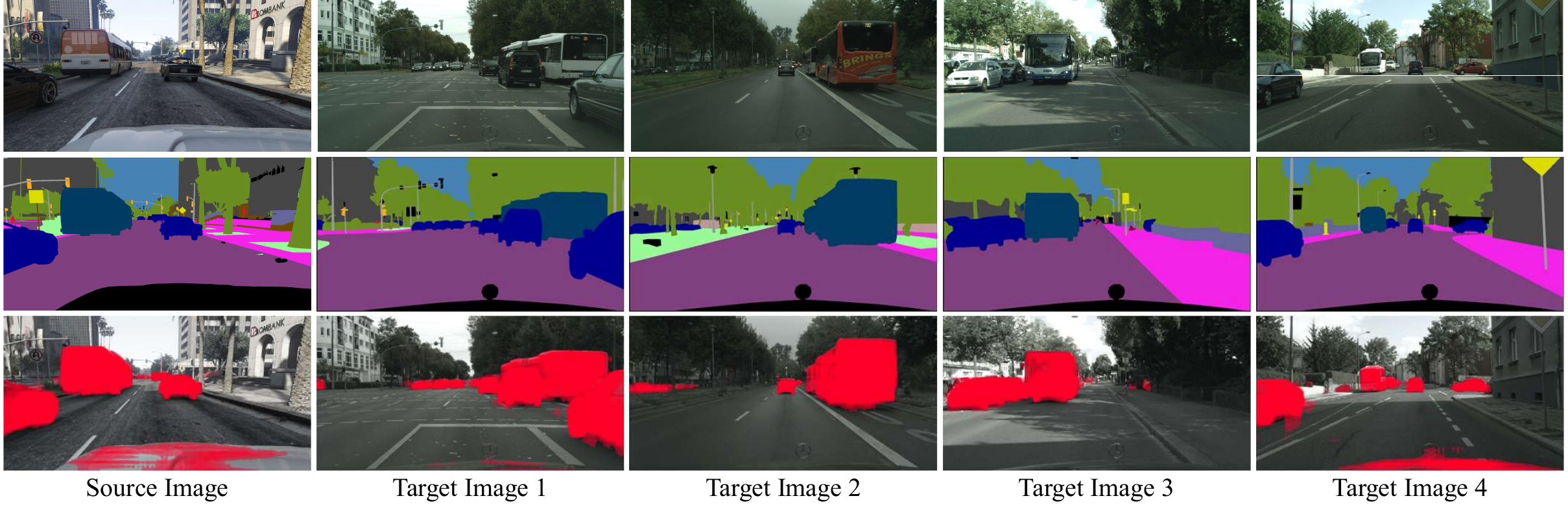}
   
     \caption{Visualization of clustered groups with randomly sampled target images. 
    (From top to bottom) Input image, ground truth, and grouping result.} 
   \label{fig:grouping_random}
\end{figure*}

	To validate the robustness of our cross-domain grouping module in diverse scenes, we further experiment our cross-domain grouping module between a source image with randomly sampled target images.
	In each row, we present the input image followed by the ground truth and grouping result.
	As shown in \figref{fig:grouping_random}, we can observe our cross-domain grouping network can cluster semantic consistent objects (e.g. bus and cars) with a diverse scene of target images.
	It verifies that our network can effectively divide the samples into semantic consistent groups with different pairs of source and target images.
	As a consequence, it is robust to scene variation while training the overall network.
	
\subsubsection{Visualization of class distribution for each group}\label{sec:3}
\begin{figure*}
   \centering
   \includegraphics[width=0.99 \linewidth]{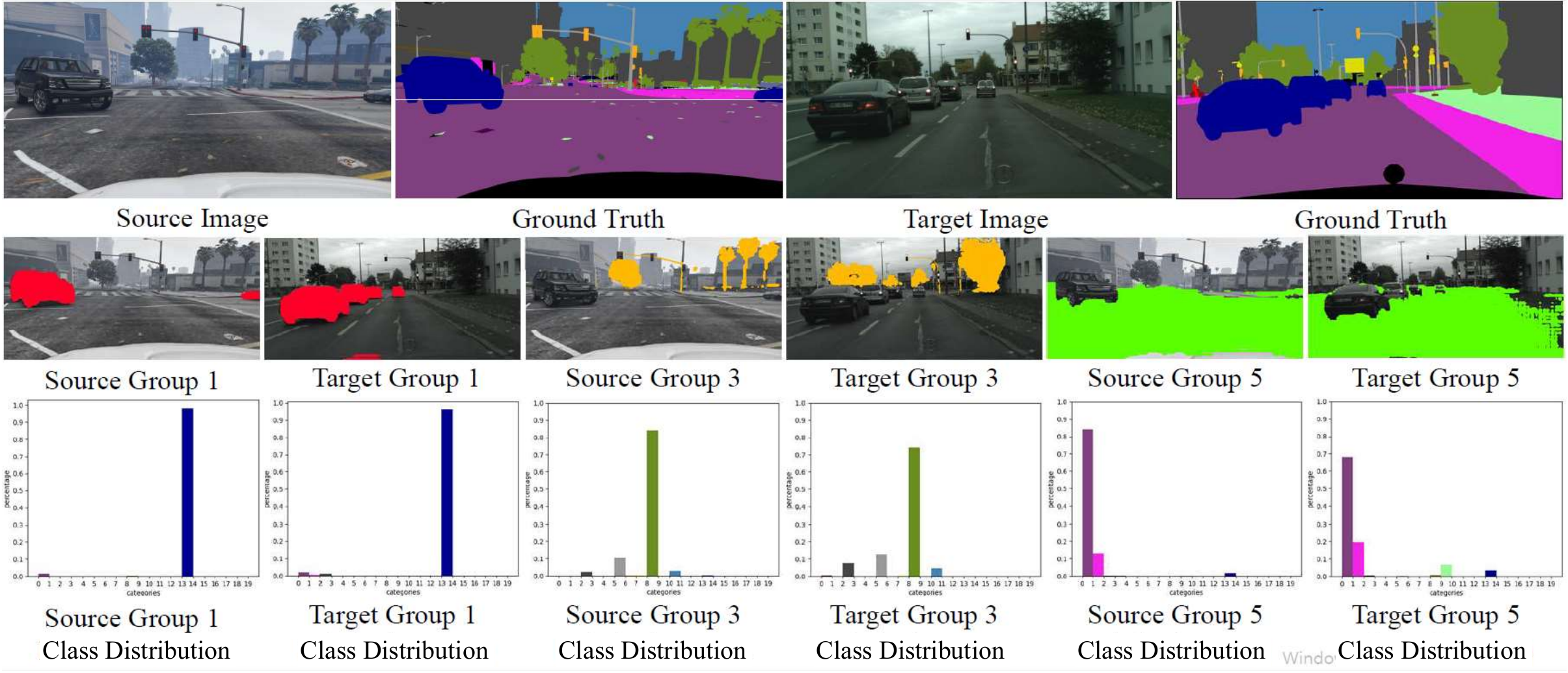}
   
     \caption{(From top to bottom) Input image, grouping result, and class distribution for each group.} 
   \label{fig:label_histogram}
\end{figure*}

    We aim to make class distribution of each group ${Q}^{k}_{l}$ between the source and target domains to be consistent, while the class distribution of different groups with the same domain to be different through losses (1) and (2).
    In \figref{fig:label_histogram}, we visualize the class distribution of the source and target domain for each group.
	We can observe the class distribution of each group is consistent between the source and target domain. For example, road and sidewalk can be grouped into the same cluster across the source and target domain.
	Also, we can see the class distribution of different groups is to be different.
    It demonstrates our cross-domain grouping network that trained with the semantic consistency and orthogonality constraints can effectively divide the multi-modal complex distribution into several simple distributions, while the class distribution of different domains in the same group can be consistent.

\subsubsection{More Results}
\begin{figure*}
   \centering
   \includegraphics[width=0.99 \linewidth]{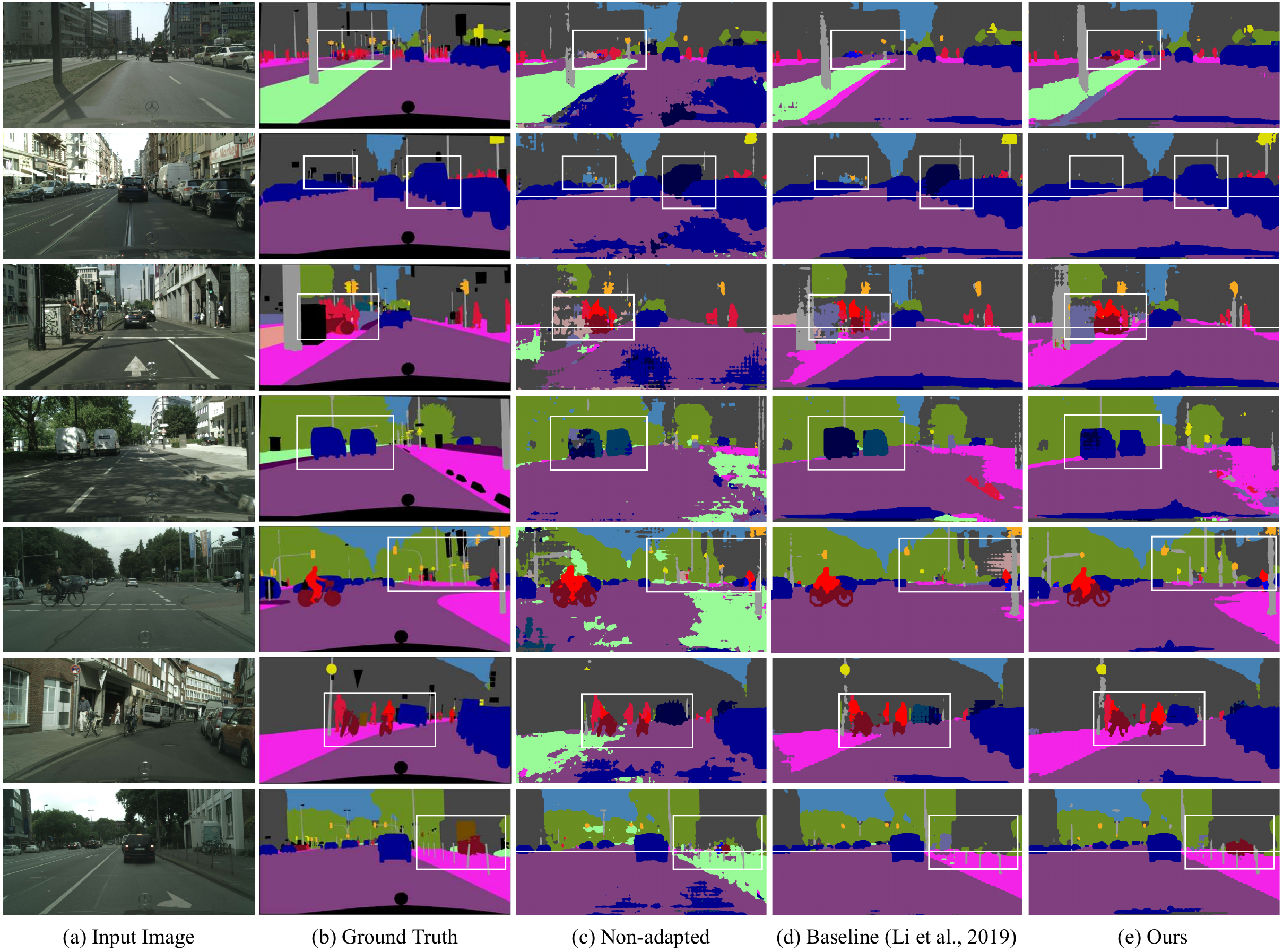}
   
     \caption{Qualitative results of domain adaptation on GTA5 $\rightarrow$ Cityscapes. 
    (From left to right) Input image, ground-truth, non-adapted result, baseline~\cite{Yunsheng19} result and ours result.} 
   \label{fig:segmentation_output}
\end{figure*}

\begin{figure*}
   \centering
   \includegraphics[width=0.99 \linewidth]{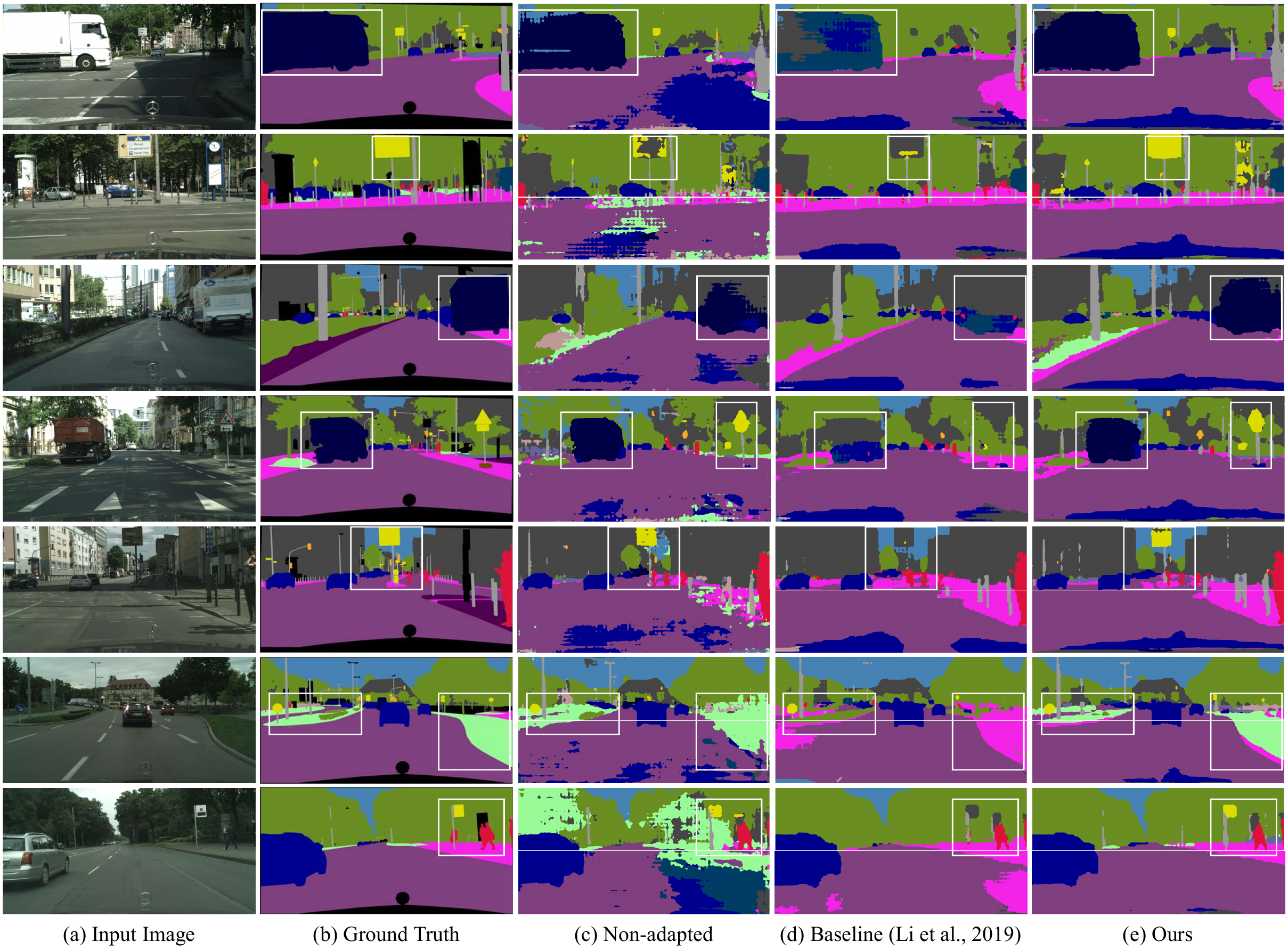}
   
     \caption{Qualitative results of domain adaptation on SYNTHIA $\rightarrow$ Cityscapes. 
    (From left to right) Input image, ground-truth, non-adapted result, baseline~\cite{Yunsheng19} result and ours result.} 
   \label{fig:synthia_output}
\end{figure*}

We provide more qualitative results for GTA5 $\rightarrow$ Cityscapes in \figref{fig:segmentation_output} and SYNTHIA $\rightarrow$ Cityscapes in \figref{fig:synthia_output}.
In each column, we present target images, ground truth, the results of the non-adapted model, baseline~\cite{Yunsheng19}, and the proposed method.
We show that our approach often yields better segmentation results using group-level alignment.
Specifically, our model achieves large performance improvement on minority classes (\eg{} pole, traffic sign, and bike) via group-level class equivalence scheme.

\bibliographystyle{aaai21}
\bibliography{egbib}